# Recognizing Handwritten Mathematical Expressions as LaTex Sequences Using a Multiscale Robust Neural Network

**Hongyu Wang [a], Guangcun Shan[*,a]**

[a] School of Instrumentation Science and Optoelectronics Engineering, Beihang University, No.37 Xueyuan Road, BeiJing, 100191, China

**\* Email: gcshan@buaa.edu.cn,**

**Abstract**

In this paper, a robust multiscale neural network is proposed to recognize handwritten mathematical expressions and output LaTeX sequences, which can effectively and correctly focus on where each step of output should be concerned and has a positive effect on analyzing the two-dimensional structure of handwritten mathematical expressions and identifying different mathematical symbols in a long expression. With the addition of visualization, the model's recognition process is shown in detail. In addition, our model achieved 49.459% and 46.062% ExpRate on the public CROHME 2014 and CROHME 2016 datasets. The present model results suggest that the state-of-the-art model has better robustness, fewer errors, and higher accuracy.

## 1. Introduction

Mathematical expressions are used in all branches of science and academic pursuits to express equalities, inequalities, and other mathematical relationships to help humans understand physical and other phenomena. Mathematical expressions are often easier to write by hand than by using computerized tools such as MathML, LaTeX, and various equation editors. Therefore, the ability for a machine to automatically recognize and encode mathematical expressions could enhance the ability of researchers, instructors, students to write mathematical expressions in computer applications.

The way people provide input information into machines has gradually changed from keyboard to handwritten input with the growing popularity of smart phones and other devices. In the case of mathematical expressions, this requires machines to recognize handwritten mathematical expressions (HMEs) as sequences that can be encoded using a markup language such as LaTeX. Compared to traditional text recognition approaches, there remain some difficulties in recognizing HMEs, making it a challenging research issue. First, the mathematical expression is a sequence with a two-dimensional (2-D) structure. Similar symbols represent different meanings depending on placement, such as "$2^x$" and "2x." Meanwhile, some LaTeX notations that must be recognized does not appear directly in mathematical expressions, but should be obtained by analyzing the 2-D structure of the expression. For example Fig.1(a) and Fig.1(b) show two HMEs that contain the implicit LaTeX notations "/sqrt," "^," and "/frac." Second, mathematical expressions cover a wide range of mathematical symbols, including arithmetic symbols, Greek letters, English letters, numbers, and so on. This requires that algorithms should have the ability to recognize different kinds of symbols. Finally, there are some confusing symbols in mathematical expressions, especially in HMEs, such as the letter "O" and the number "0."

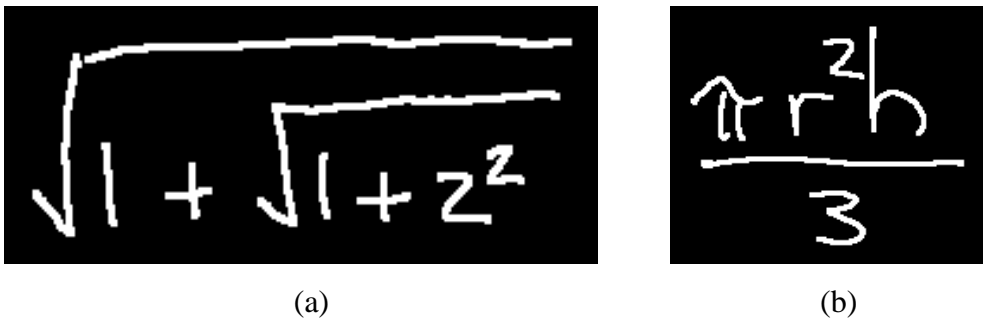

(a)                                    (b)

Fig.1: Example of implicit LaTex notations in HMEs. (a) shows "/sqrt" and "^" in expression. (b) shows "/frac" and "^" in expression.

To solve these problems, our paper proposes a "watch step-by-step" network model, which adopts an encoder-decoder framework. The encoder is used to extract features of mathematical expressions and generate coding vectors. The decoder is used to decode the coding vectors, determine the attention weights, and output prediction LaTeX sequences. The paper is organized as follows. In Section 2, a brief review of related work is presented. In Section 3, the "watch step-by-step" model based on the encoder-decoder framework is presented in detail. Experimental results and visualization of output are discussed in Section 4. Finally, Section 5 presents the conclusions.

## 2. Related Works

The beginning of the task of mathematical expression recognition dates back to 1968, as proposed by Anderson[3]. Since then, many researchers have tried to investigate this problem using different approaches. The recognition of mathematical expressions can be divided into two categories by the recognition objects: printed mathematical expression recognition and HME recognition. For the recognition of a printed mathematical expression, it is easy to identify the symbols because of its neat mathematical structure, clear writing, and clear spatial location. However, for HMEs, the recognition process may have a series of problems such as blurred writing and unclear spatial positions due to different writing habits of the writer [4-11]. Therefore, this case requires a strong, robust model.

For recognizing of printed mathematical expressions, Lavirotte and Pottier[4] proposed a model which is based on graph grammars. Kumar et al.[5] proposed an identification method consisting of three stages: symbol generation, structure analysis and code generation. Lately, Zhang et al.[12] proposed a model using Watch, Attend, and Parse to perform image recognition that includes handwritten mathematical expressions, which can avoid problems that stem from symbol segmentation and does not require a predefined expression grammar.

## 3. Framework of "Watch Step by Step"

As mentioned previously, our "watch step-by-step" network model is built on the encoder-decoder framework, the structure of which is shown in Fig.2. In Fig.2, part A shows the structure of the encoder and part B shows the structure of the decoder's attention. Then, part C is the decoder's output layers.

3.1 Dense Encoder Network

In our task, because HMEs of different lengths exist in images of different sizes, an efficient method of extracting image features is needed. Our model adopts a dense encoder network based on DenseNet[13], proposed by Gao Huang in 2017. DenseNet connects different convolution layers by using dense concatenates, so that the gradient can be transmitted without loss in the process of back propagation. At the same time, DenseNet has no significant increase in parameters compared with other deep convolution neural networks, such as GoogleNet[14] or ResNet[15], by using a smaller growth rate and dimensionality reduction of the feature map in the transition layer.

In our model, as shown in part A in Fig.2, we use three dense blocks, with 6, 12, and 24 layers. The growth rate value of each dense block is 24, which means that the dimension of every output convolution layer in a given dense block is 24. The feature maps are sent to the decoder as a context vector.

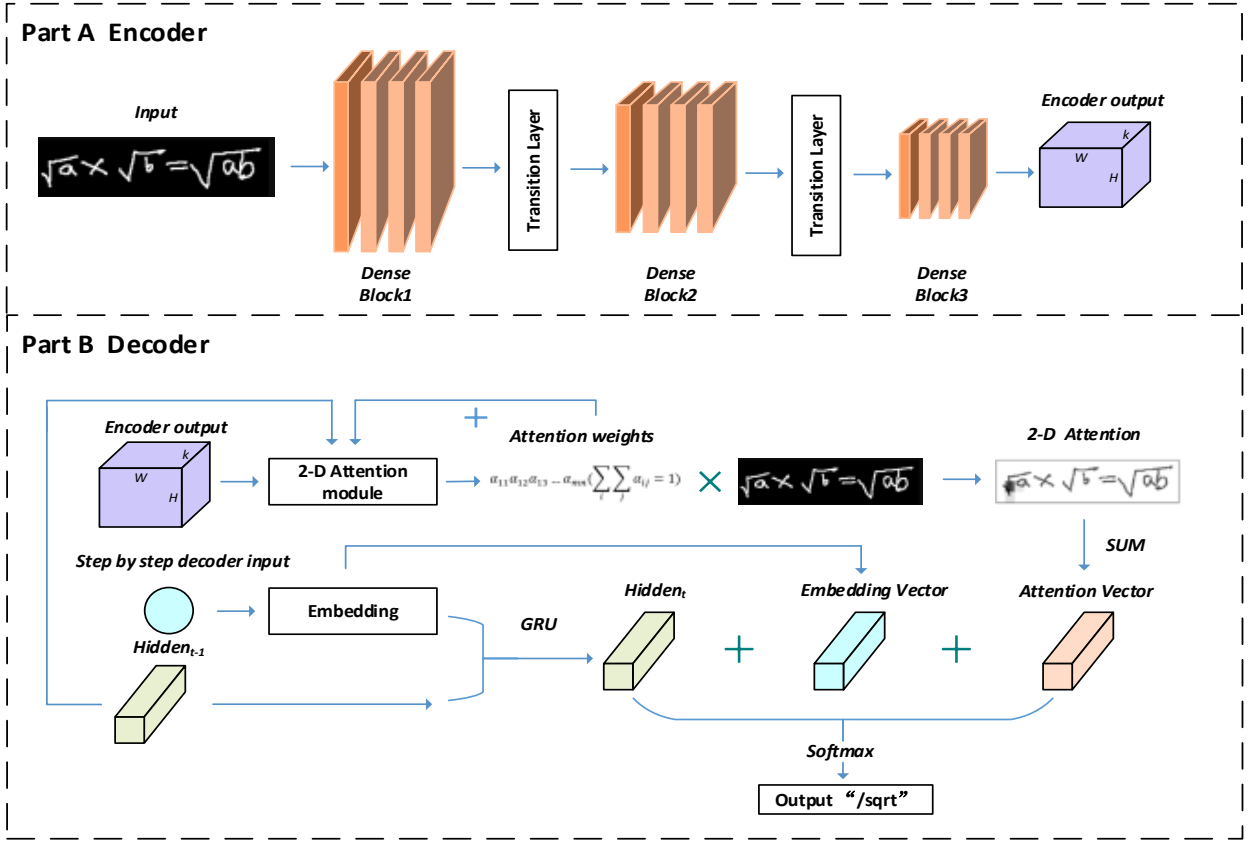

Fig.2: The Encoder-Decoder Neural Network used in our model. In this figure, the model recognizes a picture containing a HME of $\sqrt{a} \times \sqrt{b} = \sqrt{ab}$.

## 3.2 Attention-Based Decoder Network

In this model, the decoder's function is as follows: First, it generates 2-D attention weights according to the encoder's output to identify the area where the model should pay attention at present. Then, by combining input characters, the attention region, and hidden vectors, the output of the current step is determined.

(1) 2-D Attention Module

The attention model is widely used in machine translation, image understanding, and other fields. Its function is to enable the model to mimic the way human beings understand information, that is, to focus on the valuable information in the whole information set. The attention model, first proposed in 2014 [16], solved the problem of language alignment of different lengths in machine translation. Different researchers then applied the attention model to various tasks and achieved good results. For example, Xu et al. [17] proposed a visual attention-based image annotation model and also the concepts of soft attention and hard attention. Chen et al. [18] applied domain knowledge-based attention to a recommendation system.

In this study, in order to better analyze 2-D HMEs, a 2-D attention module was applied in the decoder. This 2-D attention module acts on the feature map of the encoder's output, which causes the decoder to focus on the correct position in the decoding process. 2-D attention weights are obtained from equations (1) to (3). In equation (1), F represents the feature map of the encoder's output, and $h_{t-1}$ represents the previous hidden layer vector. $f_1$ and $f_2$ represent the fully connected layers; their function is to find the matching extent between the hidden and feature map. "Conv" represents a convolution layer; its function is to expand the receptive field of the neural network to avoid weighting a small area too much. Finally, "BN" represents batch normalization [19], which can accelerate the training and avoid the gradient disappearance of the neural network. This operation yields the output $M_t$ of size $B \times K'$ $\times H \times W$. In equation (2), first, $M_t$ is activated through the tanh function. Then, we use a multi-layer perception to change the dimension of $M_t$ from $B \times K' \times H \times W$ to $B \times 1 \times$ $H \times W$. The reason for this operation is to aggregate the feature map's information of all channels into one channel. And then, we can obtain the matching extent in a 2-D plane. In equation (3), the matching extent is calculated as 2-D attention weights through the 2-D softmax function.

$$M_t = BN\{Conv[f_1(h_{t-1}) + f_2(F)]\} \qquad (1)$$

$$a^t = MLP_{k' \to 1}[\tanh(M_t)] \qquad (2)$$

$$\alpha_{ij}^t = \text{soft max}(a_{ij}^t) = \frac{\exp(a_{ij}^t)}{\sum_m \sum_n \exp(a_{mn}^t)} \qquad (3)$$

In equation(1), F represents the feature map of the encoder's output, and $h_{t-1}$ represents the previous hidden layer vector. $f_1$ and $f_2$ represent the fully connected layers; their function is to find the matching extent between the hidden and feature map. "Conv" represents a convolution layer; its function is to expand the receptive field of the neural network to avoid weighting a small area too much. Finally, "BN" represents batch normalization[19], which can accelerate the training and avoid the gradient disappearance of the neural network. This operation yields the output $M_t$ of size B×K′×H×W. In equation(2), first, $M_t$ is activated through the tanh function. Then, we use a multi-layer perception to change the dimension of $M_t$ from B×K′×H×W to B×1×H×W. The reason for this operation is to aggregate the feature map's information of all channels into one channel. Then, we can obtain the matching extent in a 2-D plane. In equation(3), the matching extent is calculated as 2-D attention weights through the 2-D softmax function.

Although the above equations have been able to correctly compute 2-D attention weights, it is not enough for HMEs. Therefore, based on the coverage model in [12,20], we design a coverage model suitable for the recognition of HMEs.

Coverage model is a method that can tell the network which parts have been paid attention to by summing up past attention weights. In this paper, we can get coverage weights given by

$$Coverage = Coverage + Conv(Att_{t-1}) \qquad (4)$$

where $Att_{t-1}$ is the attention weights at last moment and "Conv" is a convolution layer. After adding the coverage model, the structure of 2-D attention module is shown in Fig.3. The effect of coverage model will be described in detail in Section 4. Meanwhile, the equation(1) will be changed to

$$M_{tc} = BN\{Conv[f_1(h_{t-1}) + f_2(F) + f_3(C_{t-1})]\} \qquad (5)$$

where $M_{tc}$ is the output after using coverage and then it will be computed by equation(2) and (3) as described above.

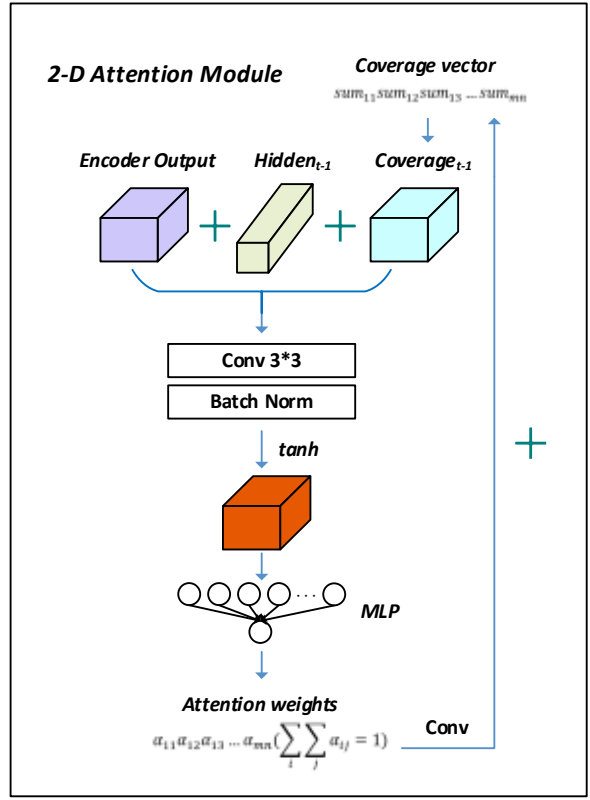

Fig.3: The structure of 2-D attention module

(2)Output LaTex characters step by step

In this work, the output of the model is determined by the embedding vector, the hidden vector, and the attention vector. Among them, the embedding vector is calculated according to

$$Emb_t = Embedding(x_t) \tag{6}$$

where $x_t$ is the input LaTex notation in current time. Embedding is a method of mapping a single word to a word vector. It is useful for expressing the difference and similarity between two words. Then, the hidden vector is computed by

$$h_t = GRU(h_{t-1}, Emb_t) \tag{7}$$

where "GRU" (Gated Recurrent Unit) is a general model, proposed in [21]. GRU is improved on the basis of LSTM (Long Short-Term Memory)[22] in order to solve the problem that recurrent neural networks cannot deal with long sequences effectively.

Next, we obtain the attention vector by

$$att_{ij}^t = \alpha_{ij}^t \times F \qquad (8)$$

$$s_t = \sum_i \sum_j att_{ij}^t \qquad (9)$$

Finally, the model sums these vectors, and then outputs the LaTeX notations in the current stage through the softmax function

$$output_t^{'} = f_{emb}(Emb_t) + f_{hid}(h_t) + f_{att}(s_t) \qquad (10)$$

$$output_t = soft\max[\mathrm{dr}opout(output_t^{'})] \qquad (11)$$

In equation(11), dropout is a necessary way to avoid overfitting[23]. In training this model, we used 0.5 as the dropout rate; we discuss the effect of dropout in Section 4.

## 4. Experiment

### 4.1 Data Set and Implementation details

The data set used in this study was the public CROHME data set, which is the largest HMEs data set available; we chose the 2014 and 2016 versions as our training data set and validated the performance of our model. In the CROHME 2014 data set, the training and testing sets comprise 8836 and 986 images, respectively, containing HMEs. In the CROHME 2016 data set, the training set includes all the data of CROHME 2014 and further extends 1147 new mathematical expressions to yield the test set for 2016. The CROHME 2014 and 2016 data sets include 110 different math symbols, including numbers, nearly all mathematical operators, and the start and stop symbols, <sos> and <eol>. In addition, these data sets contain HMEs of different lengths, from three characters to fifty characters. Thus, it is difficult for a model to correctly identify the HMEs in the images.

In the training process, Stochastic Gradient Descent (SGD)[24] + Nesterov Momentum[25] were used as the optimizer in our model. In addition, the initial learning rate was 0.0001; then, the learning rate is reduced to one-tenth of the original after 80 epochs, 120 epochs and 150 epochs. After that, we used L2-norm to relieve overfitting. The model was trained in two TITAN XP GPUs with batch size 6.

## 4.2 WER Loss and ExpRate

In the CROHME data set, we focused on two indexes of the model's recognition results: WER loss and ExpRate. WER loss focuses on the editing distance between two sequences. Editing distance is the minimum number of single-character operations required to convert one sequence into another. The operations include insert, delete, and replace, and the WER loss is computed by

$$Loss_{WER} = \frac{N_{insert} + N_{delete} + N_{replace}}{Length_{label}} \tag{12}$$

Then, when the WER Loss is zero, this prediction result is completely correct and we can get ExpRate as

$$ExpRate = \frac{N_{correct}}{N_{all}} \tag{13}$$

where $N_{correct}$ is the correct number of predictions for the model and $N_{all}$ is the number of HMEs in testing data set.

## 4.3 Recognition results and Visualization

In this section, we discuss in detail the results of our proposed model with the CROHME data set. All the experimental results shown in the tables were completed on the premise that the maximum area of the input image is 200,000 pixels and the longest mathematical expression contains 48 symbols. Under this premise, there were 925 images in the CROHME 2014 testing data set (ignoring 30 images) and 1092 images in the CROHME 2016 testing data set (ignoring 53 images).

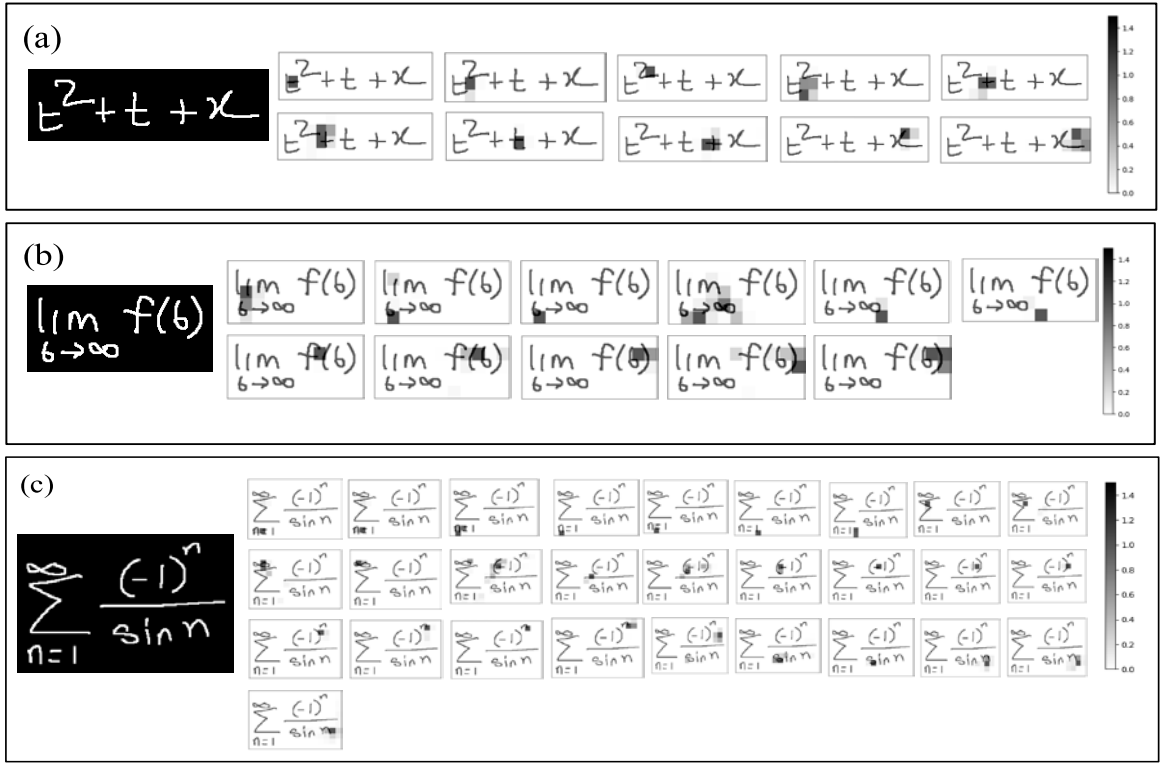

Fig.4: Visual output of the "Watch step by step" model. In picture (a), the predicted output is 't', '^', '{', '2', '}', '+', 't', '+', 'x', '<eol>' . In picture (b), the predicted output is '\\lim', '_', '{', 'b', '\\rightarrow', '\\infty', '}', 'f', '(', '6', ')', '<eol>' . In picture (c), the predicted output is '\\sum', '_', '{', 'n', '=', '1', '}', '^', '{', '\\infty', '}', '\\frac', '{', '(', '-', '1', ')', '^', '{', 'n', '}', '}', '{', '\\sin', 'n', '}', '<eol>' . Obviously, this model achieves completely correct prediction in these results.

Table 1 and Table 2 show the test results for the CROHME 2014 and 2016 data sets. It can be seen that our proposed model achieves the best results with ExpRate metric. The symbols ≤1(%), ≤2(%), and ≤3(%) in the tables mean the percentage of at most one error, two errors, and three errors between the predicted results and the ground truth. This helps to further understand the effect of the model. For example, Table 1 shows that approximately 17% of the HMEs were predicted with only one incorrect character in CROHME 2014 data set, which means that our model still has much room for improvement.

Table 1: The testing results in CROHME 2014 data set.(System I to VII can be seen in detail in [1,12]. System III is removed due to the use of additional data. '*' means the testing results in a single model by using the official open source code, so it is lower than the results in [26].)

| System | ExpRate(%) | ≤1(%) | ≤2(%) | ≤3(%) |
|---|---|---|---|---|
| I [1] | 37.22 | 44.2 | 47.3 | 50.2 |
| II [1] | 15.01 | 22.3 | 26.6 | 27.7 |
| IV [1] | 18.97 | 28.2 | 32.4 | 33.4 |
| V [1] | 18.97 | 26.4 | 30.8 | 33.0 |
| VI [1] | 25.66 | 33.2 | 36.0 | 37.3 |
| VII [1] | 26.06 | 33.9 | 38.5 | 40.0 |
| WAP$_{VGG}$ [12] | 46.55 | 61.2 | 65.2 | 66.1 |
| WAP$_{Dense}$ [26] | 41.95* | 54.7* | 65.4* | 71.3* |
| Ours | **49.46** | **66.0** | **74.4** | **80.4** |

Table 2: The testing results in CROHME 2016 data set.

| System | ExpRate(%) | ≤1(%) | ≤2(%) | ≤3(%) |
|---|---|---|---|---|
| Tokyo | 43.94 | 50.9 | 53.7 | - |
| Nantes | 13.34 | 21.0 | 28.3 | - |
| Sao Paolo | 33.39 | 43.5 | 49.2 | - |
| WAP$_{VGG}$ | 44.55 | 57.1 | 61.6 | 62.3 |
| Ours | **46.33** | **59.6** | **69.9** | **76.6** |

Table 3: The testing results of coverage in CROHME 2014 data set. 'C' represents whether coverage is used in the model.

| | C | WER(%) | ExpRate(%) |
|---|---|---|---|
| WAP$_{VGG}$ | ✗ | 28.41 | 35.09 |
| WAP$_{VGG}$ | √ | 17.73 | 46.55 |
| Ours | ✗ | 16.18 | 45.89 |
| Ours | √ | **11.99** | **49.46** |

Table 4: The test results of different ways to avoid overfitting in CROHME 2014 data set.

| | WER(%) | ExpRate(%) | Del | Ins | Rep |
|---|---|---|---|---|---|
| Origin | 24.10 | 32.22 | - | - | - |
| + dropout | 17.02 | 39.39 | 4185 | 4201 | 2409 |
| + rotation | 13.49 | 45.89 | 3690 | 3709 | 1910 |
| + L2-norm | **11.99** | **49.46** | 3511 | 3530 | 1718 |

In Fig.4, we show the identification process of our model in detail. In Fig.4(a), 4(b), 4(c), there are three pictures containing HMEs and the right side of each picture shows in detail the output of our model step-by-step. Among them, the dark region represents the area that the model pays attention to in the current time. From the results of recognition, our model predicts the correct LaTex sequence.

The effect of coverage in the model is shown in Table 3, Fig.5, and Fig.6. In Table 3 and Fig.5, the model with or without coverage obviously differs greatly in results; the model with coverage obtains better results. In Fig.6, the images on the left and right parts show step-by-step attention by a model with and without coverage, respectively. Their outputs are "\\sin," "\\theta," "<eol>," and "\\sin," "\\theta," "\\theta," "<eol>," respectively. Obviously, the model was translated "\\theta" more than once without using coverage. Therefore, coverage can prevent over-recognition in some cases.

Table 5: The testing results in CROHME 2016 data set for using pre-training parameters in CROHME 2014 data set.

| model | Pre-training | Loss$_{\text{WER}}$ | ExpRate |
|---|---|---|---|
| A | ✗ | 14.05 | 44.05 |
| B | √ | **12.93** | **46.33** |

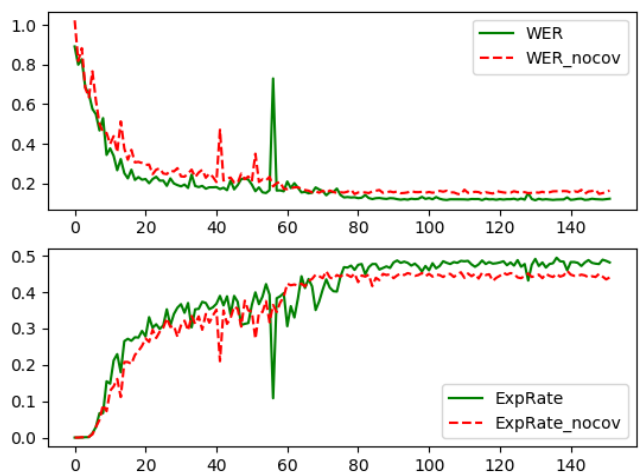

Fig.5: The WER Loss and ExpRate in the process of training, 'nocov' means a model without using coverage. The horizontal axis represents the number of epochs and the vertical axis means the WER Loss and ExpRate.

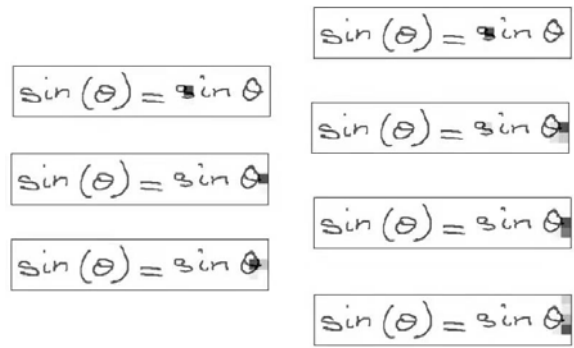

Fig.6: The images on the left and right are a step-by-step attention by a model with or without coverage. The left output is "\\sin," "\\theta," "<eol>," and the right output is "\\sin," "\\theta," "\\theta," "<eol>."

Next, the effect of regularization on the model was analyzed, because we used the fully connected layer extensively in the model. A fully connected layer is a layer with a large number of parameters, which has a strong learning ability, but it often leads to model over-fitting. From Table 4, it is clear that after adding dropout, random rotation, and L2-norm, the effect of the model was greatly improved. In this table, Del, Ins, and Rep mean the number of delete, insert, and replace operations, respectively. In Table 5, the effect of transfer learning in CROHME 2016 data set is shown. The difference between model A and model B in Table 5 is whether the pre-training parameters are used. These pre-training parameters are the parameters that achieve the best results on the CROHME 2014 data set. It is obvious that the model is initialized by pre-training parameters achieves better results and this is also validated by the positive effect of transfer learning from CROHME 2014 and the robustness. Finally, we compared the recognition results of HMEs of different lengths on the testing data set in Table 5.

## 5. Conclusion

In this paper, we propose a so-called watch step-by-step encoder-decoder model for HMEs recognition. With the encoder to extract the HMEs features and the decoder to generate LaTeX output step-by-step, our model here achieves the best results on both the CROHME 2014 and 2016 data sets. In this model, the proposed 2-D attention module with coverage are used to better analyze complex 2-D HMEs structures and avoid over-recognition. Meanwhile, the robustness of the model is greatly improved by using random image rotation, dropout, and L2-norm. Although our model has achieved better results than those of previous approaches, its ability to predict long HMEs is still insufficient. As shown in Figure 7 and Table 6, it is not able to predict long HMEs very well, which represents a highly challenging task in this research community. Therefore, in the future, we will focus on how to effectively improve the model's prediction ability for long HMEs. Besides, we would further prove the effectiveness of our proposed model on additional data sets

**Acknowledgement:**

The work was supported by the National Key R&D Program of China (Grant No. 2016YFE0204200).

**Conflict of interest:**

The authors declare no conflicts of interest.